\documentclass{article}

\usepackage[final,nonatbib]{neurips_glfrontiers_2023}

\usepackage[utf8]{inputenc} 
\usepackage[T1]{fontenc}    
\usepackage{hyperref}       
\usepackage{url}            
\usepackage{booktabs}       
\usepackage{amsfonts}       
\usepackage{nicefrac}       
\usepackage{microtype}           
\usepackage{enumitem}
\usepackage{graphicx}
\usepackage{amssymb}
\usepackage{pifont}
\usepackage{multirow}
\usepackage{rotating}
\newcommand{\cmark}{\ding{51}} 
\newcommand{\xmark}{\ding{55}}
\newcommand{\tabincell}[2]{\begin{tabular}{@{}#1@{}}#2\end{tabular}}

\title{Graph Meets LLMs: Towards Large Graph Models}

\author{Ziwei Zhang, Haoyang Li, Zeyang Zhang, Yijian Qin, Xin Wang, Wenwu Zhu\\
  Department of Computer Science and Technology, Tsinghua University. Beijing, China, 100084\\
  \texttt{\{zwzhang,xin\_wang,wwzhu\}@tsinghua.edu.cn} \\ 
  \texttt{\{lihy18,zy-zhang20,qinyj19\}@mails.tsinghua.edu.cn} \\
}

\begin{document}
\maketitle
\begin{abstract}
Large models have emerged as the most recent groundbreaking achievements in artificial intelligence, and particularly machine learning. However, when it comes to graphs, large models have not achieved the same level of success as in other fields, such as natural language processing and computer vision. In order to promote applying large models for graphs forward, we present a perspective paper to discuss the challenges and opportunities associated with developing large graph models\footnote{We maintain a curated paper list at \url{https://github.com/THUMNLab/awesome-large-graph-model}.}. First, we discuss the desired characteristics of large graph models. Then, we present detailed discussions from three key perspectives: representation basis, graph data, and graph models. In each category, we provide a brief overview of recent advances and highlight the remaining challenges together with our visions. Finally, we discuss valuable applications of large graph models. We believe this perspective can encourage further investigations into large graph models, ultimately pushing us one step closer towards artificial general intelligence (AGI). We are the first to comprehensively study large graph models, to the best of our knowledge.
\end{abstract}
\section{Introduction}
In recent years, there has been a growing interesting in large models for both research and practical applications. Large models have been particularly revolutionary in fields such as natural language processing (NLP)~\cite{GPT3,kenton2019bert,zhao2023survey} and computer vision (CV)~\cite{SAM,CLIP,rombach2022high}, where pre-training extremely large models on large-scale unlabeled data has yielded significant breakthroughs. However, graphs, which are commonly used to represent relationships between entities in various domains such as social networks, molecule graphs, and transportation networks, have not yet seen the same level of success with large models as other domains. In this paper, we present a perspective about the challenges and opportunities associated with developing large graph models. First, we introduce large graph models and outline four key desired characteristics, including graph models with scaling laws, graph foundation model, in-context graph understanding and processing abilities, and versatile graph reasoning capabilities. Then, we offer detailed perspectives from three aspects: (1) For graph representation basis, we discuss graph domains and transferability, as well as the alignment of graphs with natural languages. Our key takeaway is the significance of identifying a suitable and unified representation basis that spans diverse graph domains, which serves as a fundamental step towards constructing effective large graph models; (2) For graph data, we summarize and compare the existing graph datasets with other domains, and highlight that the availability of more large-scale high-quality graph data is resource-intensive yet indispensable; (3) For models, we systematically discuss backbone architectures, including graph neural networks and graph Transformers, as well as pre-training and post-processing techniques, such as prompting, parameter-efficient fine-tuning, and model compression. We also discuss LLMs as graph models, which is a newly trending direction. Finally, we discuss the significant impact that large graph models can have on various graph applications, including recommendation systems, knowledge graphs, molecules, finance, code and program, and urban computing and transportation. We hope that our paper can inspire further research into large graph models\footnote{There are also works to use graphs to improve large language models, such as enhancing their reasoning ability~\cite{sun2023thinkongraph,yao2023beyond,yao2023thinking,cao2023enhancing,lei2023boosting,besta2023graph,wen2023mindmap} or using graphs as tools~\cite{zhang2023toolformer,jiang2023structgpt}, which is beyond the scope of this paper.}. 

\section{Desired Characteristics of Large Graph Models}\label{sec:characters}
\begin{figure}[t]
    \centering
    \includegraphics[width = \textwidth]{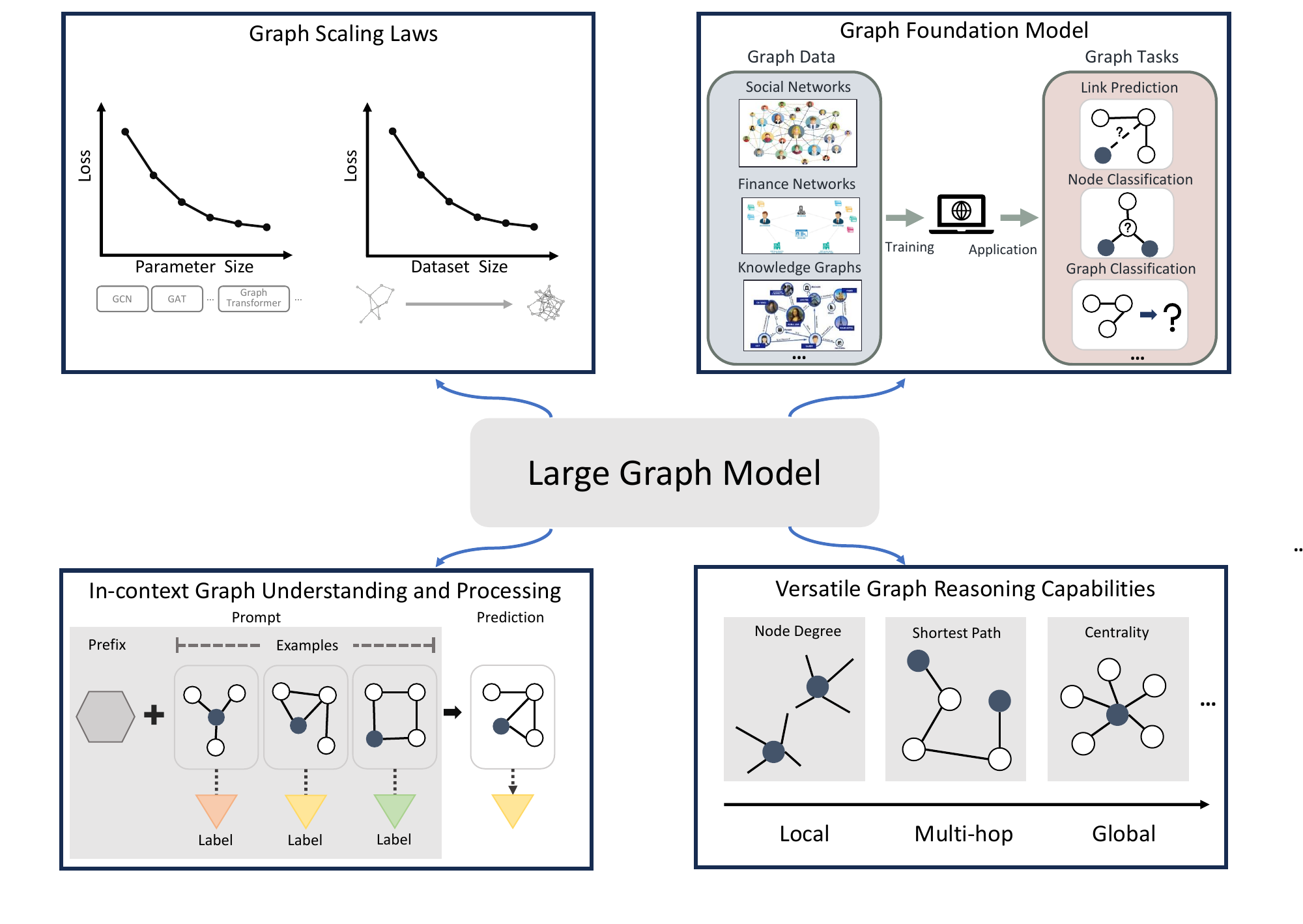}
    \caption{An illustration of desired characteristics of a large graph model.}
    \label{fig:character}
\end{figure}

Similar to large language models (LLMs)~\cite{zhao2023survey}, a large graph model can be characterized as a graph model with a vast number of parameters which empower it with abilities that are substantially more powerful than smaller models, thereby promoting the understanding, analyses, and processing of graph-related tasks. Apart from having numerous parameters, we summarize the key desired characteristics of an ideal large graph model from the following perspectives. An illustration of these characteristics is provided in Figure~\ref{fig:character}.
\begin{enumerate}[leftmargin=0.3cm]
     \item \textbf{Graph models with scaling laws}: The scaling laws indicate an empirical phenomenon where the performance of LLMs continues to improve as the model size, dataset size, and training computation increase~\cite{kaplan2020scaling}. This phenomenon offers a clear direction for enhancing performance and empowering the model to capture complex patterns and relationships within graph data. By emulating the success of LLMs~\cite{wei2022emergent}, a large graph model is expected to exhibit emergent abilities that smaller models lack. However, accomplishing this objective in large graph models is highly non-trivial, with difficulties span from collecting more graph data to solving technical problems such as addressing the over-smoothing and over-squashing problem of graph neural networks, along with engineering and system challenges.
    \item \textbf{Graph foundation model}: A large graph model holds greater value when it can serve as a graph foundation model, i.e., capable of handling different graph tasks across various domains. This requires the model to gain understandings of the inherent structural information and properties of graphs to be equipped with ``commonsense knowledge'' of graphs. The graph pre-training paradigm is a highly promising path to develop graph foundation models, as it can expose the model to large-scale unlabeled graph data and reduce the reliance on expensive and laborious collection of graph labels. Besides, a generative pre-training paradigm can potentially empower the model with the ability to generate graphs, thereby opening up possibilities for valuable applications like drug synthesis, code modeling, and network evolution analysis~\cite{guo2023generation}. It is worthy clarifying that, since graphs serve as general data representations with extreme diversity, it is exceedingly challenging, if not unlikely, to develop a ``universal graph model'' for all graph domains. Therefore, multiple graph foundation models may be necessary for different ``clusters of domains'', which is somewhat different from LLMs or foundation models in computer vision.
    \item \textbf{In-context graph understanding and processing abilities}: An effective large graph model is expected to comprehend graph contexts, including nodes, edges, subgraphs, and entire graphs, and process novel graph datasets and tasks during testing with minimum samples as well as without intensive model modifications or changes in the paradigm. This characteristic is also closed related to and can facilitate capabilities of few-shot/zero-shot graph learning~\cite{zhang2022few}, multi-task graph learning~\cite{ju2022multi}, and graph out-of-distribution generalization~\cite{li2022out}. Moreover, these abilities are vital when the input graph data and task are different between the training and testing stages. In-context learning abilities can enable large graph models to leverage knowledge learned in the pre-training stage and quickly adapt to the testing stage with desired performance.  
    \item \textbf{Versatile graph reasoning capabilities}: Although graphs span diverse domains, there exist common and fundamental graph tasks. We generally refer to  handling of these tasks as ``graph reasoning''. While there is no clear consensus on what these tasks are, some representative examples are provided as follows. Firstly, a large graph model should understand basic topological graph properties, such as graph sizes, node degrees, node connectivity, etc. These properties form the foundation for a deeper understanding of graph structures. Secondly, a large graph model should be able to flexibly and explicitly reason over multi-hop neighborhoods, enabling it to perform more sophisticated tasks. Such capabilities, akin to the chain-of-thought of LLMs~\cite{wei2022chain} in principle, can also enhance transparency in the graph decision-making process and improve model explainability~\cite{yuan2022explainability}. Lastly, besides local information, a large graph model should be able to understand and handle graph tasks that involve global properties and patterns, such as the centrality and position of nodes, overall properties of graphs, the evolution laws of dynamic graphs, etc.
\end{enumerate}

\section{Graph Representation Basis}
\subsection{Graph Domains and Transferability}
Large models, LLMs, serve as foundation models~\cite{bommasani2022opportunities}, as they can be adapted to a wide range of downstream tasks after being pre-trained. The remarkable ability of LLMs stems from the underlying assumption of the existence of a common representation basis for various NLP tasks. For instance, word tokens for natural language processing are universal and information-preserving data representations that do not rely on specific tasks. In contrast, graphs are general data structures that span a multitude of domains. Therefore, the raw input data, i.e., nodes and edges, may not always be the most suitable representation basis for handling all graph data. Nodes and edges in social networks, molecule graphs, and knowledge graphs, for instance, have distinct meanings with their unique feature and topological space. Thus, directly sharing information and transferring knowledge based on input graph data often poses significant challenges.  

It is widely believed that there exist more high-level or abstract common graph patterns, which can be shared across different graphs and tasks within a certain domain. For example, many human interpretable patterns have been identified in classical network science~\cite{newman2018networks}, such as homophily, small-world phenomenon, power-law distribution of node degrees, etc. Nevertheless, even with these high-level shared knowledge, creating effective large models that can perform well across diverse graph domains is still non-trivial. 

\subsection{Aligning with Natural Languages}
Another key competency of recent large models is their ability to interact with humans and follow instructions, as we are naturally capable of understanding languages and visual perceptions. In contrast, humans are less capable of handling graphs, especially more complex reasoning problems. As a result, communicating and instructing large models to behave for graph tasks the way we desire, especially using natural languages, is particularly challenging. We summarize three categories of strategies worth exploring to overcome this obstacle. 

The first strategy is to align the representation basis of graphs and text through a large amount of paired data, similar to computer vision in principle. If successful, we will be able to interact with graph models using natural languages. For example, we can ask the model to generate molecule graphs with desired properties or ask the model to perform challenging graph reasoning tasks. Some initial attempts have been made for text-attributed graphs~\cite{he2023explanations,duan2023simteg}, which serve as a good starting point. However, collecting such data for general graphs is much more costly and challenging than image-text pairs. 

The second strategy is to transform graphs into natural languages, and then work solely in the language basis. Some initial attempts using this strategy have been developed, where graph structures are transformed into text representations, such as the adjacency list or the edge list, and inserted into LLMs as prompts. Then, natural languages are used to perform graph analytical tasks. We provide more detailed discussions in Section~\ref{sec:LLMforGraph}. However, directly transforming graph data and tasks into languages may lose the inner structure and inductive bias for graphs, resulting in unsatisfactory task performance. More delicate designs, such as effective prompts to convert graph structures and tasks into texts, are required to further advance this strategy.   

The last category is to find other representation basis as a middle ground for different graph tasks and natural languages. The most straight-forward way is to use some hidden space of neural networks. However, it faces the challenge that deep neural networks are largely not explainable at the moment, not to mention that finding the desired shared hidden space can be frustratingly challenging. On the other hand, although humans are not capable of directly handling graph data, we can design appropriate algorithms to solve graph tasks, including many well-known algorithms in graph theory such as finding shortest paths, dynamic programming, etc. Therefore, if we can align the behavior of graph models with these algorithms, we can understand and control the behaviors of these models to a certain extent. Some efforts have been devoted in this direction, known as algorithmic reasoning~\cite{velivckovic2021neural}, which we believe contains rich potentials.

In summary, finding the suitable representation basis, potentially aligning with natural languages, and unifying various graph tasks across different domains is one fundamental step towards building successful large graph models.

\section{Graph Data}
The success of big models is largely dependent on the availability of high-quality, large-scale datasets. For instance, GPT-3 was pre-trained on a corpus of approximately 500 billion tokens~\cite{GPT3}, while CLIP, a representative model that bridges natural language processing and computer vision, was trained on 400 million image-text pairs. It is reasonable to assume that even more data has been utilized in more recent large models, such as GPT-4~\cite{openai2023gpt4}. This massive amount of data for NLP and CV tasks is typically sourced from publicly accessible human-generated content, such as web pages in CommonCrawl or user-posted photos in social media, which are easily collected from the web.

In contrast, large-scale graph data is not as easily accessible. There are typically two scenarios for graph data: numerous small-scale graphs, such as molecules, or a single/few large-scale graphs, such as social networks or citation graphs. For example, Open Graph Benchmark~\cite{hu2020open}, one of the most representative public benchmarks for graph machine learning, includes two large graph datasets: MAG240M, which contains a large citation graph with approximately 240 million nodes and 1.3 billion edges, and PCQM4M, which contains approximately 4 million molecules. However, their scale is considerably lower than the datasets used in NLP or CV. If we treat each node in MAG240M as a token (though a node may contain arguably more information) or each graph in PCQM4M as an image, these graph datasets are at least $10^3$ to $10^4$ times smaller than their NLP or CV counterparts.

In addition to the data utilized for pre-training, commonly accepted and widely adopted benchmarks, such as SuperGLUE~\cite{wang2019superglue} and BIG-bench~\cite{srivastava2023beyond} for NLP and ImageNet~\cite{deng2009imagenet} for CV, have been found to be beneficial in the development of large models. These benchmarks are  especially useful in assessing model quality and determining the most promising technical routes during the early stages. Although there are numerous benchmarks available for graph learning, such as Open Graph Benchmark~\cite{hu2020open} and Benchmarking GNN~\cite{dwivedi2023benchmarking}, it is likely that their scope, including factors like scale, task and domain diversity, and evaluation protocols, may not be suitable or sufficient for evaluating large graph models. Therefore, the creation of more specialized benchmarks can further facilitate the progress of large graph models.

Next, we summarize some principles that are helpful while collecting more graph data.
\begin{itemize}[leftmargin = 0.4cm]
\item Domain diversity: To enable large graph models handle different graph applications, it is crucial to expose the model to different domains of interests, so that large graph models can be adopted across various fields and serve as the graph foundation model.
\item Type diversity: Graphs have rich types, including homogeneous and heterogeneous, homophily and heterophily, static and dynamic, directed and undirected, weighted and unweighted, signed and unsigned, etc. The diversity of graph type is also important to empower the large graph model handle diverse downstream graphs.
\item Statistics diversity: Graphs also have varying statistics, e.g., size, density, degree distribution, etc. Such diversity should be considered to ensure the effectiveness of large graph model.
\item Task diversity: Graph tasks are also distinct, ranging from node-level, edge-level to graph-level, and from discriminative tasks such as classification and prediction to generative tasks such as graph generation. Increasing the task diversity in pre-training or post-processing phase can help developing and evaluating effective large graph models. 
\item Modality diversity: Graphs, as general data representations, can also combine different modalities of data, such as text, images, and tabular data, which can further enrich the utility of the large graph model.
\end{itemize}
In summary, the availability of high-quality graph data is critical to the development of large graph models, which requires more resources and efforts. Since collecting such graph data is difficult and costly, community-wide collaboration may be essential to accelerate this process.

\section{Graph Models}\label{sec:model}
In this section, we continue the discussion from the graph model aspect. Similar to large models in other domains, we divide our discussion into three topics: backbone architecture, pre-training, and post-processing. We also discuss LLMs as graph models, which is a recently trending direction.
\subsection{Backbone Architecture}
To date, Transformers~\cite{vaswani2017attention} have been the de facto standards for NLP and CV. However, no similar consensus has been reached for the graph domain. We briefly discuss two promising deep learning architectures for graphs: graph neural networks (GNNs) and graph transformers. 

GNNs are the most popular deep learning architectures for graphs~\cite{zhang2020deep} and have been extensively studied. Most representative GNNs adopt a message-passing paradigm, where nodes exchange messages with their neighbors to update their representations. GNNs can incorporate both structural information and semantic information such as node and edge attributes in an end-to-end manner. However, despite achieving considerable successes in many graph tasks, one key obstacle for further advancing GNNs into large models is their limited model capacity. As opposed to the scaling law in large models~\cite{kaplan2020scaling}, the performance of GNNs saturates or even dramatically drops as the model size grows. Many research efforts have been devoted to explain this problem, such as over-smoothing~\cite{rusch2023survey} and over-squashing~\cite{toppingunderstanding}, as well as strategies to alleviate it. Nevertheless, progress has not been groundbreaking. To date, most successful GNNs only have at most millions of parameters, and further scaling to billions of parameters leads to minimum or no additional improvement.

Graph Transformer is another architecture that extends and adapts the typical Transformers for graph data~\cite{min2022transformer}. In a nutshell, since classical Transformers cannot naturally process graph structures, Graph Transformer adopts various structure-encoding strategies to add graph structures to the input of Transformers~\cite{zhang2022autogt}. Graph Transformers evaluate the importance of each neighboring node, giving larger weights to nodes that provide more pertinent information. The self-attention mechanism empowers Graph Transformers the ability to dynamically adapt. One of the most successful graph Transformers is Graphormer~\cite{ying2021transformers}, which ranked first in the PCQM4M molecule property prediction task of OGB Large-Scale Challenge~\cite{hu2021ogb} in 2021. More efforts further improve Graph Transformer from various aspects including architecture designs, efficiency, model expressiveness, etc. For example, Structure-Aware Transformer (SAT)~\cite{chen2022structure} proposes a new self-attention mechanism to capture the structural similarity between nodes more effectively. AutoGT~\cite{zhang2022autogt} proposes a unified graph transformer formulation for existing graph transformer architectures and enhances the model performance using AutoML. To improve efficiency, General, Powerful, and Scalable graph Transformer (GPS)~\cite{rampavsek2022recipe} introduces a general framework with linear complexity by decoupling the local edge aggregation from the fully-connected Transformer. NAGphormer~\cite{chen2023nagphormer} also aims to address the complexity challenge of graph Transformers for large graphs by treating different hops of neighbors as a sequence of token vectors. For the expressiveness, SEG-WL test~\cite{zhu2023structural} introduces a graph isomorphism test algorithm, which can be used for assessing the structural discriminative power of graph Transformers. FeTA~\cite{bastos2022expressive} analyzes the expressiveness of graph Transformers in the spectral domain and proposes to perform attention on the entire graph spectrum.

We briefly summarize the key differences between GNNs and graph Transformers, while more discussions for the relationships between transformers and GNNs can be found ~\cite{joshi2020transformers, velivckovic2023everything, muller2023attending,kim2022pure}:
\begin{itemize}[leftmargin=0.4cm]
    \item Aggregation vs. Attention: GNNs employ message passing functions to aggregate information from neighboring nodes, whereas Graph Transformers weigh contributions from neighbors using self-attentions, potentially enhancing the flexibility for large graph models.
    \item Modeling structures: GNNs naturally incorporate graph structures in the message passing functions as an inductive bias, while graph Transformers adopt pre-processing strategies, such as structure-encoding, to incorporate structures. 
    \item Depth and Over-smoothing: As aforementioned, deep GNNs may suffer from over-smoothing, leading to a decrease in their discriminative power. Graph Transformers, on the other hand, do not exhibit similar issues empirically. One plausible explanation is that Graph Transformers adaptively focus on more relevant nodes, enabling them to effectively filter and capture informative patterns.
    \item Scalability and Efficiency: GNNs, with their relatively simpler operations, may offer computational benefits for certain tasks. In contrast, the self-attention mechanism between node pairs in Graph Transformers can be computationally intensive, especially for large graphs. Considerable efforts have been dedicated to further enhancing the scalability and efficiency for both methods.
\end{itemize}
While both GNNs and Graph Transformers have made remarkable progress, it is not very clear which one, or some other architectures, may be best suited as the backbone for large graph models. Besides empirical evidence from trials and errors, further research into how large models work and what graph problems they may solve could bring principled advancements. It is also worth noting that most graph tasks relate to reasoning rather than perception. Therefore, the inductive bias in architecture designs usually does not come from mimicking human brains. 

In our opinion, given the scale of existing graph datasets, GNNs are still a strong backbone model thanks to their strong inductive bias and expressive power. However, as the size of the training graph datasets continues to increase, graph Transformers may become more powerful through increasing the number of parameters and gradually become the prevailing approach.

\subsection{Pre-training}\label{sec:pretraining}
Pre-training, as a widely adopted practice in NLP with well-known models like BERT~\cite{kenton2019bert} and GPT~\cite{radford2018improving}, involves training a model on a massive dataset before applying it for specific tasks. The primary objective is to capture general patterns or knowledge present in the data and subsequently adapt the pre-trained model to meet downstream requirements. Graph pre-training, also known as unsupervised or self-supervised graph learning, has received significant attention in recent years~\cite{liu2022graph,wu2021self}. It aims to capture the inherent structural patterns within the training graph data, analogous to how language models capture the syntax and semantics of languages. As explained in Section~\ref{sec:characters}, we recognize pre-training as an essential paradigm for large graph models. Next, we provide a more detailed discussion of graph pre-training. 

Compared to the straightforward yet effective masking operation used in language modeling, graph pre-training strategies are more diverse and complicated, ranging from contrastive to predictive/generative approaches. Generally, graph pre-training methods leverage the rich structural and semantic information in the graph to introduce pretext learning tasks. Through these tasks, the pre-trained model learns useful node, edge, or graph-level representations without relying on explicitly annotated labels. In contrastive pre-training methods, positive and negative graph samples are constructed through various graph data augmentation techniques, followed by optimizing contrastive objectives, such as maximizing the mutual information between positive and negative pairs. On the other hand, in generative and predictive methods, specific components of the graph data, such as node features and edges, are first hide by masking. Then, the graph model aims to reconstruct the masked portions, which serve as pseudo-labels for pre-training. For more details, we refer readers to dedicated surveys~\cite{liu2022graph,wu2021self}. 

We summarize the desired benefits of graph pre-training using the following ``four-E" principle:
\begin{itemize}[leftmargin=0.4cm]
    \item \textbf{E}ncoding structural information: Unlike pre-training methods for other types of data, such as languages and images, which focus primarily on semantic information, graphs contain rich structural information. Pre-training large graph models essentially needs to integrate structural and semantic information from diverse graph datasets. This also highlights the unique challenges and opportunities of graph pre-training. 
    \item \textbf{E}asing data sparsity and label scarcity: Large graph models, with their substantial model capacity, are prone to overfitting when confronted with specific tasks that have limited labeled data. Pre-training on a wide range of graph datasets and tasks can act as a regularizing mechanism, preventing the model from overfitting to a specific task and improving generalization performance.
    \item \textbf{E}xpanding applicability domains: One of the hallmarks of pre-training is the ability to transfer learned knowledge across various domains. By pre-training large graph models on diverse graph datasets, they should be able to capture a wide range of structural patterns, which can then be applied, adapted, or fine-tuned to graph data in similar domains, maximizing the model's utility.
    \item \textbf{E}nhancing robustness and generalization. Pre-training methods can expose large graph models to diverse graphs with distinct characteristics, including varying sizes, structures, and complexities. This exposure can potentially lead to more robust models that are less sensitive to adversarial perturbations~\cite{hendrycks2019using}. Moreover, models trained in this manner are more likely to generalize well to unseen graph data or novel graph tasks.
\end{itemize}
In summary, graph pre-training is not merely a beneficial or supplementary step, but a pivotal and necessary paradigm for large graph models. 

\subsection{Post-processing}\label{sec:postprocess}
After obtaining a substantial amount of knowledge through pre-training, LLMs still require post-processing to enhance their adaptability to downstream tasks. Representative post-processing techniques include prompting~\cite{liu2023pre}, parameter-efficient fine-tuning~\cite{ding2023parameter}, reinforcement learning with human feedbacks~\cite{ouyang2022training}, and model compression~\cite{zhu2023survey}. For graphs, some recent efforts have also been devoted to study post-processing techniques for pre-trained models.

Prompting originally refers to methods that provide specific instructions to language models for generating desired contents for downstream tasks. Recently, constructing prompts with an in-context learning template demonstrates great effectiveness in LLMs~\cite{dong2022survey}. Language prompts usually contain a task description and a few examples to illustrate the downstream tasks. Graph prompts, which mimic natural language prompts to enhance downstream task performance with limited labels and enable interaction with the model to extract valuable knowledge, have been extensively studied~\cite{liu2023graphprompt}. One significant challenge for graph prompts is the unification of diverse graph tasks, spanning from node-level and link-level to graph-level tasks. In contrast, tasks in natural language can be easily unified as language modeling under specific constraints. To tackle this challenge, GPPT~\cite{sun2022gppt} unifies graph tasks into edge prediction, considering that a typical node classification task can be reformulated as the link prediction task between the structure-token and the task-token. Each structure-token represents a node in the graph data, and each task-token corresponds to a class. GraphPrompt~\cite{liu2023graphprompt} further extends the idea and unifies link prediction, node classification, and graph classification as subgraph similarity calculation by describing node and graph classes as prototypical subgraphs. Similarly, ProG~\cite{sun2023all} reformulates node and edge-level tasks as graph-level tasks and further proposes multi-task prompting by realizing prompting as a learnable token that is directly added to the node feature, mirroring the prefix phrase prompting technique in NLP. ProG also employs meta learning to learn prompting for different tasks. Other graph prompts such as PRODIGY~\cite{huang2023prodigy}, GPF~\cite{fang2023universal}, Gare~\cite{jing2023deep}, SGL-PT~\cite{zhu2023sglpt}, DeepGPT~\cite{shirkavand2023deep}, G-Prompt~\cite{huang2023promptbased}, CPP~\cite{zhu2023chain}, KGTransformer~\cite{zhang2023KGTransformer}, SAP~\cite{ge2023enhancing}, HetGPT~\cite{ma2023hetgpt}, ULTRA-DP~\cite{chen2023ultradp}, PGCL~\cite{gong2023prompt}, and TAG~\cite{li2023promptbased} follow similar principles.

Parameter-efficient fine-tuning refers to techniques where only a small portions of model parameters are optimized, while the rest is kept fixed. Besides reducing computational costs, it also helps to enable the model to adapt to new tasks without forgetting the knowledge obtained in pre-training, preserving the general capabilities of the model while allowing for task-specific adaptation. Graph parameter-efficient fine-tuning has also recently begun to received attention. For example, AdapterGNN~\cite{li2023adaptergnn} and G-Adapter~\cite{gui2023g} both investigate adapter-based fine-tuning techniques for graph models, aiming to reduce the number of tuneable parameters while preserving comparable accuracy. Specifically, AdapterGNN tunes GNNs by incorporating two adapters, one inserted before and another one after the message passing process. On the other hand, G-Adapter focuses on graph transformers and introduces a message passing process within the adapter to better utilize graph structural information. S2PGNN~\cite{wang2023search} further proposes to search for architecture modifications to improve the adaptivity of the fine-tuning stage. 

Model compression aims to reduce the memory and computational demands of models through various techniques, including knowledge distillation, pruning, and quantization, which are particularly valuable when deploying large models in resource-constrained environments. Here, we focus on quantization, which has gained popularity and proven effectiveness in LLMs~\cite{zhao2023survey}, and refer readers to dedicated surveys for other methods~\cite{tian2023knowledge,zhang2023survey,sergi2021survey}. Quantization entails reducing the precision of numerical values used by the model while preserving model performance to the greatest extent possible. In the case of large models, post-training quantization (PTQ) is particularly preferred, as it does not require retraining. PTQ in graph learning has also been explored in SGQuant~\cite{feng2020sgquant}, which proposes a multi-granularity quantization technique that operates at various levels, including graph topology, layers, and components within a layer. Other methods such as Degree-Quant~\cite{tailor2020degree}, BiFeat~\cite{ma2022bifeat}, Tango~\cite{chen2023tango}, VQGraph~\cite{yang2023vqgraph}, $A^2Q$~\cite{zhu2022rm}, and AdaQP~\cite{wan2023adaptive} adopt a quantization-aware training scheme, which are inspiring but cannot be used standalone during the post-processing stage.

In summary, the success of post-processing techniques shown in LLMs has sparked interest in similar research in the graph domain. However, due to the unavailability of large graph models at present, the assessment of these methods is limited to relatively small models. Therefore, it is crucial to further verify their effectiveness when applied to large graph models, and more research challenges and opportunities may arise.

\subsection{LLMs as Graph Models}\label{sec:LLMforGraph}
Recent research has also explored the potential of directly utilizing LLMs for solving graph tasks. The essential idea is to transform graph data, including both graph structures and features, as well as graph tasks, into natural language representations, thereby treating graph problems as regular NLP problems. In the following discussion, we first provide a brief overview of representative methods, and then provide detailed comparisons of different methods.

NLGraph~\cite{wang2023can} conducts a systematic evaluation of LLMs, such as GPT-3 and GPT-4, on eight graph reasoning tasks in natural language, spanning varying levels of complexity, including connectivity, shortest path, maximum flow, simulating GNNs, etc. It empirically finds that LLMs show preliminary graph reasoning abilities, but struggle with more complex graph problems, potentially because they solely capture spurious correlations within the problem settings. Meanwhile, GPT4Graph~\cite{guo2023gpt4graph} also conducts extensive experiments to evaluate the graph understanding capabilities of LLMs across ten distinct tasks, such as graph size and degree detection, neighbor and attribute retrieval, etc. It reveals the limitations of LLMs in graph reasoning and emphasizes the necessity of enhancing their structural understanding capabilities. LLMtoGraph~\cite{liu2023evaluating} also tests GPT-3.5 and GPT-4 for various graph tasks and makes some interesting observations. 

More recently, Graph-LLM~\cite{chen2023exploring} systematically investigates the utilization of LLMs in text-attributed graph through two strategies: LLMs-as-Enhancers, where LLMs enhance the representations of text attributes of nodes before passing them to GNNs, and LLMs-as-Predictors, where LLMs are directly employed as predictors. Comprehensive studies have been conducted on these two pipelines across various settings, and the empirical results provide valuable insights into further leveraging LLMs for graph machine learning. InstructGLM~\cite{ye2023natural} further introduces scalable prompts designed to describe the graph structures and features for LLM instruction tuning, which enables tuned LLMs to perform various graph tasks during the inference stage in a generative manner. Experiments conducted on GNN benchmarks empirically show the strong potential of adopting LLMs for graph machine learning. 

Next, we summarize and compare different models related to LLMs as graph models. The overall summarization is shown in Table~\ref{tab:LLMasGraph}. Specifically, we category the key features into three groups: model architectures, modeling Graph structure for LLMs, and graph data.

For model architectures, we summarize the following designs:
\begin{itemize}[leftmargin = 0.4cm]
    \item Final Predictor: whether the model utilizes GNNs or LLMs to get the final prediction. 
    \item LLMs: which LLMs are utilized in the model. Typical examples include GPT-3~\cite{GPT3}, GPT-4~\cite{openai2023gpt4}, Llama 2~\cite{touvron2023llama}, etc. 
    \item Need Fine-tuning: whether the model needs to be fine-tuned. Note that if the model does not necessarily require fine-tuning, but could be fine-tuned to further improve the performance, we mark it as no. Note that close-sourced LLMs such as GPT-3 and GPT-4 are not tunable.
    \item Tunable Components: which parts of the model can be fine-tuned, such as GNNs, graph Transformers, and LLMs.
    \item Receptive Field: how many hops of neighbors can be perceived when making predictions. K-hop indicates the receptive field is determined by the architectures, e.g., the number of layers in GNNs. 
\end{itemize}

One key challenge of using LLMs as graph models is to model graph structures and inject them into LLMs. As this is usually achieved through prompts, we make the following summarization: 
\begin{itemize}[leftmargin = 0.4cm]
    \item Prompt Type: whether the model uses textual prompts (i.e., descriptions in texts) or neural prompts (e.g., through hidden layers in neural networks).  
    \item Prompt Details: the details of the prompt. Common textual prompts include adjacency lists and neighborhood descriptions, and typical neural prompts include GNNs and graph Transformers. 
    \item Advanced graph-specific prompt: the types of advanced graph-specific prompts, if they are proposed in the paper.  
\end{itemize}
Lastly, we summarize the graph data used in the experiments. Note that we focus on the experiments conducted in the original papers, but extensions are possible, e.g., handling larger graphs by using more computational resources or applying the model to other tasks through minor modifications. 
\begin{itemize}[leftmargin = 0.4cm]
    \item Dataset Type: what type of graphs are utilized in the experiments, including synthetic graphs, TAGs, knowledge graphs (KGs), and general graphs. 
    \item Tasks: what type of tasks are considered in the experiments, including algorithmic tasks (various from degree counting to finding shortest paths, etc.), node classification, link prediction, question answering (QA), etc. 
    \item \#Nodes: the approximate number of nodes handled by the model. If sampling is applied, we only count the sampled nodes. 
    \item Node Feature: whether and what type of node features can be utilized in the model, including no attributes, text attributes, and general attributes. 
\end{itemize}	

\begin{sidewaystable}
\caption{A summarization of different models related to LLMs as graph models.}\label{tab:LLMasGraph}
\tabcolsep=0.05cm
\resizebox{0.95\linewidth}{!}{
\begin{tabular}{|c|ccccc|ccc|cccc|} \hline
\multirow{3}{*}{Method}                                             & \multicolumn{5}{c|}{Architecture}                                                                              & \multicolumn{3}{c|}{Modeling Graph Structure for LLMs}                                                                                                      & \multicolumn{4}{c|}{Graph Data}                                                                                                                           \\ \cline{2-13}
                                                                    &  \tabincell{c}{Final \\Predictor} & LLMs                         &  \tabincell{c}{Need\\Finetune}         & \tabincell{c}{Tunable\\Components} & \tabincell{c}{Receptive\\Field} & 
                                                                    \tabincell{c}{Prompt\\Type} & 
                                                                    \tabincell{c}{Prompt Details}                                     & 
                                                                    \tabincell{c}{Advanced \\Graph-specific Prompt}                                                          & 
                                                                    \tabincell{c}{Dataset\\Type}         & Tasks                                                                                                   & \# Nodes & 
                                                                    \tabincell{c}{Node\\Feature} \\ \hline
NLGraph~\cite{wang2023can}                    & LLM             & GPT-3/4                      & \xmark & -                  & Full            & Text        & Edge list                                          &  \tabincell{c}{Build-a-Graph Prompting\\ Algorithmic Prompting} & Synthetic            & Algorithm                                                                                               & $10^2$   & No           \\ \hline
GPT4Graph~\cite{guo2023gpt4graph}             & LLM             & GPT-3                        & \xmark & -                  & 2-hop           & Text        & \tabincell{c}{Graph modelling language \\ graph markup language} & -                                                                                       & Synthetic, KG        & Algorithm, QA                                                                                           & $10^2$   & Text         \\ \hline
Graph-LLM(predictor)~\cite{chen2023exploring} & LLM             & GPT-3                        & \xmark & -                  & 2-hop           & Text        & Neighbor description                               & Neighbor Summary                                                                        & TAG                  & Node classification                                                                                     & $10^2$   & Text         \\ \hline
Graph-LLM(enhancer)~\cite{chen2023exploring}  & GNN             & GPT-3, E5, etc.              & \cmark & GNN/LLM            & K-hop           & \multicolumn{2}{c}{-}                                            & -                                                                                       & TAG                  & Node classification                                                                                     & $10^6$   & General      \\ \hline
LLMtoGraph~\cite{liu2023evaluating}           & LLM             & GPT-3/4, Vicuna, etc.         & \xmark & -                  & Full            & Text        & Edge list                                          & -                                                                                       & Synthetic            & Algorithm                                                                                               & $10^2$   & No           \\ \hline
InstructGLM~\cite{ye2023natural}              & LLM             & T5, LLama-2                  & \cmark & LLM                & 2-hop           & Text        & Neighbor description                               & -                                                                                       & TAG                  & Node classification                                                                                     & $10^5$   & Text         \\ \hline
LLM-Structured-Data~\cite{huang2023llms}      & LLM             & GPT-3                        & \xmark & -                  & 2-hop           & Text        & Neighbor description                               & Attention extraction/prediction                                                     & TAG                  & Node classification                                                                                     & $10^3$   & Text         \\ \hline
OneForAll~\cite{liu2023all}                   & GNN             & GPT-3,Bert                   & \cmark & GNN                & K-hop           & Text        & -                                                  & -                                                                                       & TAG,KG               &  \tabincell{c}{Node classification\\ Link Prediction\\ Graph Classification} & $10^5$   & Text         \\ \hline
GraphText~\cite{zhao2023graphtext}            & LLM             & GPT-3, Llama-2               & \xmark & LLM                & 2-hop           & Text        & Graph syntax tree + GNN                            & Graph Syntax Tree                                                                       & General              & Node classification                                                                                     & $10^3$   & General      \\ \hline
GraphQA~\cite{fatemi2023talk}                 & LLM             & PaLM-2                       & \xmark & LLM                & Full            & Text        & Task-specific relation encoding                    & -                                                         & Synthetic            & Algorithm                                                                                               & $10^2$   & No           \\ \hline
Hu et al.~\cite{hu2023text}                   & LLM             & GPT-3/4                      & \xmark & -                  & 1-hop           & Text        & Neighbor description                               & -                                                                                       & TAG                  &  \tabincell{c}{Node classification\\ Link Prediction\\ Graph Classification}  & $10^5$   & Text         \\ \hline
GraphLLM~\cite{chai2023graphllm}              & LLM             & LLama-2                      & \cmark & GT  & Full            & Neural      & Graph Transformer                                  & -                                                                                       & Synthetic            & Algorithm                                                                                               & $10^2$   & Text         \\ \hline
LLM-GNN~\cite{chen2023labelfree}              & GNN             & GPT-3                        & \cmark & GNN                & K-hop           & \multicolumn{2}{c}{-}                                            & -                                                                                       & TAG                  & Node classification                                                                                     & $10^6$   & General      \\ \hline
ENG~\cite{yu2023empower}                      & GNN             & GPT-3, Llama-2               & \cmark & GNN                & K-hop           & Text        & Neighbor description                               & -                                                                                       & General              & Node classification                                                                                     & $10^5$   & General      \\ \hline
Wenkel et al. ~\cite{wenkel2023pretrained}    & LLM             & GPT-2/3                      & \xmark & LLM                & Full            & Text        & Edge list                                          & -                                                                                       & \tabincell{c}{Synthetic\\Molecules} & Graph classification                                                                                    & $10^2$   & Text         \\ \hline
GraphGPT~\cite{tang2023graphgpt}              & LLM             & Llama-2, Vicuna              & \cmark & GNN                & 2-hop           & Neural      & GNN + Instruction tuning                           & -                                                                                       & TAG                  & \tabincell{c}{Node classification\\ Link Prediction\\ Graph Matching}       & $10^5$   & General      \\ \hline
LLM4DyG~\cite{zhang2023llm4dyg}               & LLM             & GPT-3, Llama-2, etc. & \xmark & -                  & Full            & Text        & Temporal edge list                                 & \tabincell{c}{Disentangled \\ Spatial-Temporal Thoughts}                                                  & Synthetic            & Dynamic Algorithm                                                                                       & $10^2$   & No           \\ \hline
DGTL~\cite{qin2023disentangled}               & LLM             & LLama-2                      & \cmark & GNN                & K-hop           & Neural      & Disentagled GNN                                    & -                                                                                       & TAG                  & Node classification                                                                                     & $10^4$   & Text         \\ \hline
Stechly et al.~\cite{stechly2023gpt4}         & LLM             & GPT-4                        & \xmark & -                  & Full            & Text        & Edge list                                          & Iterative prompting                                                                     & Synthetic            & Graph Coloring                                                                                          & $10^2$   & No           \\ \hline
SimTeG~\cite{duan2023simteg}                  & GNN             & E5,Roberta                   & \cmark & GNN                & K-hop           & \multicolumn{2}{c}{-}     & -                                                                                                                                 & TAG                  & \tabincell{c}{Node classification\\ Link Prediction}                          & $10^6$   & Text         \\ \hline
GNP~\cite{tian2023graph}                      & LLM             & T5                           & \cmark & GNN/LLM            & K-hop           & Neural      & GNN + subgraph retrival                            & -                                                                                       & KG                   & QA                                                                                                      & $10^4$   & Text         \\ \hline
RoG~\cite{luo2023reasoning}                   & LLM             & GPT-3/ Llama-2               & \cmark & LLM                & L-hop           & Text        & Retriveal reasoning                                & -                                                                                       & KG                   & QA                                                                                                      & $10^4$   & Text         \\ \hline
TAPE~\cite{he2023explanations}                & GNN             & GPT-3                        & \cmark & GNN/LLM            & K-hop           & \multicolumn{2}{c}{-}                                            & -                                                                                       & TAG                  & Node classification                                                                                     & $10^5$   & Text       \\  \hline
\end{tabular}
}
\end{sidewaystable}

Although still in their early stages, these works highlight that LLMs also represent a promising avenue for developing large graph models, which is worthy further exploration and investigation.

\subsection{Summary}
To summarize, substantial research efforts have been devoted to studying various aspects of graph models. However, there is currently no clear framework for effectively integrating these techniques into large graph models. Consequently, more efforts are required to compare existing methods and develop advanced models. In this endeavor, automated graph machine learning techniques~\cite{zhang2021automated}, such as graph neural architecture search, can be valuable in reducing human effort and accelerating the trial-and-error process.

\section{Applications}
Instead of attempting to overwhelmingly handle various graph domains and tasks, it may be more effective to focus on specific graph-related vertical fields by leveraging domain knowledge and domain-specific datasets. In this section, we highlight several graph application scenarios that can significantly benefit from large graph models.

\subsection{Recommendation System}
Graph data naturally exists in recommendation systems. For example, the interaction between users and items can be modeled as a bipartite graph or more complex heterogeneous graphs that include clicks, buys, reviews, and more. Currently, LLMs for recommendation systems focus on modeling semantic information~\cite{wu2023survey}, while explicitly utilizing the structural information of graphs has the potential to yield better results~\cite{wu2022graph}. A potential challenge is that graphs in  recommendation system are usually multi-modal~\cite{tao2020mgat}, covering text, images, interactions, etc. Since large models for multi-modal data are not yet mature, significant efforts are needed to develop truly effective large graph models for recommendation systems.

\subsection{Knowledge Graph}
Knowledge graphs are widely adopted to store and utilize ubiquitous knowledge in human society. LLMs have been used for various knowledge graph tasks~\cite{pan2023unifying,ye2023natural}, including construction, completion, and question answering. Despite their achievements, most of these methods focus primarily on the textual information, leaving the structural and relational information of knowledge graphs under-explored. Large graph models, potentially combined with existing LLMs, can greatly complement the status quo and further promote research and application of knowledge graphs.

\subsection{Molecules}
Graphs are natural representations for molecules, where nodes represent atoms and edges indicate bonds. Building effective graph models for molecules can advance various applications, including molecular property prediction and molecular dynamics simulations, ultimately benefiting drug discovery. Currently, some variants of LLMs are applied to molecules~\cite{bagal2021molgpt,qian2023large} by first transforming molecules into strings using SMILES~\cite{weininger1988smiles}, which allows molecules to be represented and generated as regular texts. Nevertheless, graphs serve as a more natural way to represent the structural information of molecules with numerous modeling advantages~\cite{wieder2020compact}. Meanwhile, a great number of graph-based pre-training techniques have also been developed for molecules~\cite{xia2022systematic}, including multi-modal strategies~\cite{liu2023gitmol}. Besides, molecule data is relatively easier to collect, e.g., ZINC20~\cite{irwin2020zinc20} contains millions of purchasable compounds. Therefore, we believe graph-based or graph-enhanced large models for molecule modeling can soon to be expected.

\subsection{Finance}
Graph machine learning has proven to be beneficial for multiple financial tasks such as stock movement prediction and loan risk prediction~\cite{wang2022review}. Moreover, the large abundance of financial data makes it possible to construct domain-specific large models, exemplified by BloombergGPT~\cite{wu2023bloomberggpt}. By combining the strengths of both worlds, the application of large graph models in the field of finance holds great promise. A potential challenge lies in the sensitive and private nature of most financial data, making industries reluctant to release related models and data to the public. Efforts are required to promote open-source initiatives and democratization~\cite{liu2023fingpt,yang2023fingpt} to fully unleash the potential of large graph models in the finance area. 

\subsection{Code and Program}
Thanks to the large amount of code data available on repository hosting platforms such as GitHub, LLMs show remarkable ability in understanding and generating codes and programs. Notable examples include CodeX~\cite{chen2021evaluating}, AlphaCode~\cite{yujia2022code}, and GPT-4~\cite{openai2023gpt4}, which have exerted a significant impact on the programming landscape, potentially even reshaping it. In addition to treating codes and programs as textual data, graphs offer a natural means to represent the structural aspects of codes. For example, abstract syntax trees, including control flow graph, data flow graph, etc., effectively capture the syntactic structure of source codes~\cite{allamanis2018learning}. Studies have demonstrated that the integration of graphs can further enhance the performance of LLMs by providing complementary information~\cite{guo2021graphcodebert}. Therefore, large graph models hold valuable potential for a wide range of code and program-related tasks, including code completion and generation, code search, code review, program analysis and testing, among others.

\subsection{Urban Computing and Transportation}  
Graph data is pervasive in the domains of urban computing and transportation, such as road networks. Therefore, graph machine learning can benefit many applications, including traffic forecasting, various urban planning and management tasks, crime prediction, and epidemic control~\cite{rahmani2023transportation,jin2023spatio}. Moreover, large-scale urban data naturally exists, such as mobility data collected from GPS and diverse sensors. Currently, some LLM-based large models have been explored for urban computing and transportation, such as TransGPT~\cite{transGPT}. Nevertheless, their focus has primarily revolved around natural language related applications, leaving developing large graph models for broader and more comprehensive utilization still an open opportunity. One major technical challenge in the process lies in that graph data in urban and transportation contexts is dynamic in nature, containing complicated spatial-temporal patterns. Thus, a large graph model needs to effectively capture both structural and temporal information to achieve satisfactory performance.

\subsection{Beyond}
The application scenarios we have outlined above are by no means exhaustive. Considering that graph machine learning has been widely adopted across diverse domains ranging from industrial applications, such as fault diagnosis~\cite{chen2021graph}, IoT~\cite{dong2023graph}, power systems~\cite{liao2021review}, and time-series analysis~\cite{jin2023survey}, to AI for science~\cite{wang2023scientific}, such as physics~\cite{shlomi2020graph,dezoort2023graph}, combinatorial optimization~\cite{cappart2023combinatorial}, material science~\cite{reiser2022graph}, and neural science~\cite{bessadok2022graph}, exploring the usage of large graph models holds extremely rich potentials.  

\section{Conclusion}
In summary, large graph models can potentially revolutionize the field of graph machine learning, but they also give rise to a multitude of challenges, ranging from the representation basis, graph data, graph models, and applications. Meanwhile, promising endeavors are being undertaken to tackle these challenges, creating exciting opportunities for both researchers and practitioners. We hope that our perspective will inspire continued efforts and advancements for large graph models.
\medskip
\section*{Acknowledgement}
This work was supported by the National Key Research and Development Program of China No. 2020AAA0106300, National Natural Science Foundation of China (No. 62222209, 62250008, 62102222, 62206149), Beijing National Research Center for Information Science and Technology under Grant No. BNR2023RC01003, BNR2023TD03006, China National Postdoctoral Program for Innovative Talents No. BX20220185, China Postdoctoral Science Foundation No. 2022M711813, and Beijing Key Lab of Networked Multimedia. All opinions, findings, conclusions, and recommendations in this paper are those of the authors and do not necessarily reflect the views of the funding agencies. Xin Wang and Wenwu Zhu are corresponding authors.

\bibliographystyle{unsrt}
\bibliography{neurips_glfrontiers_2023_arXiv}

\end{document}